\documentclass[11pt,a4paper]{article}
\usepackage[hyperref]{emnlp2020}
\usepackage{times}
\usepackage{url}
\usepackage{latexsym}
\usepackage{bm}
\usepackage{booktabs}
\usepackage{multirow}
\usepackage{arydshln}
\usepackage{amsmath}

\usepackage{graphicx} 

\usepackage{todonotes}

\newcommand{\Ni}{({\em i})~}
\newcommand{\Nii}{({\em ii})~}

\usepackage{microtype}

\aclfinalcopy 

\makeatletter
\def\adl@drawiv#1#2#3{%
        \hskip.5\tabcolsep
        \xleaders#3{#2.5\@tempdimb #1{1}#2.5\@tempdimb}%
                #2\z@ plus1fil minus1fil\relax
        \hskip.5\tabcolsep}
\newcommand{\cdashlinelr}[1]{%
  \noalign{\vskip\aboverulesep
           \global\let\@dashdrawstore\adl@draw
           \global\let\adl@draw\adl@drawiv}
  \cdashline{#1}
  \noalign{\global\let\adl@draw\@dashdrawstore
           \vskip\belowrulesep}}
\makeatother

\title{We Can Detect Your Bias:\\ Predicting the Political Ideology of News Articles}

\author{
    Ramy Baly$^1$,
    Giovanni Da San Martino$^2$,
    James Glass$^1$,
    Preslav Nakov$^2$\\
    
    $^1$MIT Computer Science and Artificial Intelligence Laboratory\\
    $^3$Qatar Computing Research Institute, HBKU\\

    \texttt{\{baly,glass\}@mit.edu}\\
    \texttt{\{gmartino,pnakov\}@hbku.edu.qa}
}

\date{}

\begin{document}

\maketitle

\begin{abstract}
We explore the task of predicting the leading political ideology or bias of news articles. First, we collect and release a large dataset of 34,737 articles that were manually annotated for political ideology --left, center, or right--, which is well-balanced across both topics and media. We further use a challenging experimental setup where the test examples come from media that were not seen during training, which prevents the model from learning to detect the source of the target news article instead of predicting its political ideology. From a modeling perspective, we propose an adversarial media adaptation, as well as a specially adapted triplet loss. We further add background information about the source, and we show that it is quite helpful for improving article-level prediction. Our experimental results show very sizable improvements over using state-of-the-art pre-trained Transformers in this challenging setup.
\end{abstract}

\section{Introduction}

In any piece of news, there is a chance that the viewpoint of its authors and of the media organization they work for, would be reflected in the way the story is being told. The emergence of the Web and of social media has lead to the proliferation of information sources, whose leading political ideology or bias may not be explicit. Yet, systematic exposure to such bias may foster intolerance as well as ideological segregation, and ultimately it could affect voting behavior, depending on the degree and the direction of the media bias, and on the voters' reliance on such media~\cite{dellavigna2007fox,iyengar2009red,10.1145/2505515.2505623,graber2017mass}. 
Thus, making the general public aware, e.g.,~by tracking and exposing bias in the news is important for a healthy public debate given the important role media play in a democratic society.

Media bias can come in many different forms, e.g.,~by omission, by over-reporting on a topic, by cherry-picking the facts, or by using propaganda techniques such as appealing to emotions, prejudices, fears, etc. \cite{EMNLP2019:propaganda:finegrained,DaSanMartinoSemeval20task11,IJCAI2020:propaganda:survey}
Bias can occur with respect to a specific topic, e.g.,~COVID-19, immigration, climate change, gun control, etc. \cite{ICWSM2020:Unsupervised:Stance:Twitter,stefanov-etal-2020-predicting}
It could also be more systematic, as part of a political ideology, which in the Western political system is typically defined as left vs. center vs. right political leaning.

Predicting the bias of individual news articles can be useful in a number of scenarios.
For news media, it could be an important element of internal quality assurance as well as of internal or external monitoring for regulatory compliance.
For news aggregator applications, such as Google News, it could enable balanced search, similarly to what is found on AllSides.\footnote{\url{http://allsides.com/}}
For journalists, it could enable news exploration from a left/center/right angle.
It could also be an important building block in a system that detects bias at the level of entire news media~\cite{baly2018predicting,baly2019multi,baly2020written}, such as the need to offer explainability, i.e.,~if a website is classified as left-leaning, the system should be able to pinpoint specific articles that support this decision.

In this paper, we focus on predicting the bias of news articles as left-, center-, or right-leaning.
Previous work has focused on doing so at the level of news media~\cite{baly2020written} or social media users \cite{ICWSM2020:Unsupervised:Stance:Twitter}, but rarely at the article level~\cite{kulkarni2018multi}.
The scarce article-level research has typically used distant supervision, assuming that all articles from a given medium should share its overall bias, which is not always the case.
Here, we revisit this assumption.

Our contributions can be summarized as follows:

\begin{itemize}
    \item We create a new dataset for predicting the political ideology of news articles. 
    The dataset is annotated at the article level and covers a wide variety of topics, providing balanced left/center/right perspectives for each topic.
    \item We develop a framework that discourages the learning algorithm from modeling the source instead of focusing on detecting bias in the article.
    We validate this framework in an experimental setup where the test articles come from media that were not seen at training time.
    We show that adversarial media adaptation is quite helpful in that respect, and we further propose to use a triplet loss, which shows sizable improvements over state-of-the-art pre-trained Transformers.
    \item We further incorporate media-level representation to provide background information about the source, and we show that this information is quite helpful for improving the article-level prediction even further.
\end{itemize}

The rest of this paper is organized as follows:
We discuss related work in Section~\ref{sec:related_work}. Then, we introduce our dataset in Section~\ref{sec:dataset}, we describe our models for predicting the political ideology of a news article in Section~\ref{sec:methodology}, and we present our experiments and we discuss the results in Section~\ref{sec:results}.
Finally, we conclude with possible directions for future work in Section~\ref{sec:conclusion}.


\section{Related Work}
\label{sec:related_work}

Most existing datasets for predicting the political ideology at the news article level were created by crawling the RSS feeds of news websites with known political bias~\cite{kulkarni2018multi}, and then projecting the bias label from a website to all articles crawled from it, which is a form of distant supervision. The crawling could be also done using text search APIs rather than RSS feeds~\cite{horne2019different,gruppi2020nelagt2019}. 

The media-level annotation of political leaning is typically obtained from specialized online platforms, such as News Guard,\footnote{\url{http://www.newsguardtech.com}} AllSides,\footnote{\url{http://allsides.com/}} and Media Bias/Fact Check,\footnote{\url{http://mediabiasfactcheck.com}} where highly qualified journalists use carefully designed guidelines to make the judgments.

As manual annotation at the article level is very time-consuming, requires domain expertise, and it could be also subjective, such annotations are rarely available at the article level. As a result, automating systems for political bias detection have opted for using distant supervision as an easy way to obtain large datasets, which are needed to train contemporary deep learning models.

Distant supervision is a popular technique for annotating datasets for related text classification tasks, such as detecting hyper-partisanship~\cite{Horne:2018:ANL:3184558.3186987,DBLP:journals/corr/PotthastKRBS17} and propaganda/satire/hoaxes~\cite{Rashkin}.
For example, \citet{kiesel-etal-2019-semeval} created a large corpus for detecting hyper-partisanship (i.e.,~articles with extreme left/right bias) consisting of 754,000 articles, annotated via distant supervision, and additional 1,273 manually annotated articles, part of which was used as a test set for the SemEval-2019 task~4 on Hyper-partisan News Detection. 
The winning system was an ensemble of character-level CNNs~\cite{jiang2019team}.
Interestingly, all top-performing systems in the task achieved their best results when training on the manually annotated articles only and ignoring the articles that were labeled using distant supervision, which illustrates the dangers of relying on distant supervision.

\citet{BARRONCEDENO20191849} extensively discussed the limitations of distant supervision in a text classification task about article-level propaganda detection, in a setup that is similar to what we deal with in this paper: the learning systems may learn to model the source of the article instead of solving the task they are actually trained for. 
Indeed, they have shown that the error rate may drastically increase if such systems are tested on articles from sources that were never seen during training, and that this effect is positively correlated with the representation power of the learning model.
They analyzed a number of representations and machine learning models, showing which ones tend to overfit more, but, unlike our work here, they fell short of recommending a practical solution. 

\citet{budak2016fair} measured the bias at the article level using crowd-sourcing. This is risky as public awareness of media bias is limited~\cite{elejalde2018nature}. Moreover, the annotation setup does not scale. Finally, their dataset is not freely available, and their approach of randomly crawling articles does not ensure that topics and events are covered from different political perspectives.

\citet{lin-etal-2006-side} built a dataset annotated with the ideology of 594 articles related to the Israeli-Palestinian conflict published on \url{bitterlemons.org}. The articles were written by two editors and 200 guests, which minimizes the risk of modeling the author style.
However, the dataset is too small to train modern deep learning approaches. 

\citet{kulkarni2018multi} built a dataset using distant supervision and labels from AllSides. Distant supervision is fine for the purpose of training, but they also used it for testing, which can be problematic. Moreover, their training and test sets contain articles from the same media, and thus models could easily learn to predict the article's source rather than its bias. In their models, they used both the text and the URL contents of the articles.

Overall, political bias has been studied at the level of news outlet~\cite{DBLP:conf/interspeech/Dinkov0KN19,baly2018predicting,baly2020written,zhang-etal-2019-tanbih}, user~\cite{ICWSM2020:Unsupervised:Stance:Twitter}, article~\cite{DBLP:journals/corr/PotthastKRBS17,saleh2019team}, and sentence~\cite{sim-etal-2013-measuring,10.1145/2505515.2505623}.
In particular, \citet{baly2018predicting} developed a system to predict the political bias and the factuality of news media.
In a follow-up work,~\citet{baly2019multi} showed that bias and factuality of reporting should be predicted jointly.
A finer-grained analysis is performed in~\cite{Horne:2018:ANL:3184558.3186987}, where a model was trained on 10K sentences from a dataset of reviews~\cite{pang-lee-2004-sentimental}, and used to discriminate objective versus non-objective sentences in news articles. 
\citet{lin-etal-2006-side} presented a sentence-level classifier, where the labels were projected from the document level.

\begin{figure*}[ht]
    \centering
    \includegraphics[width=1\textwidth]{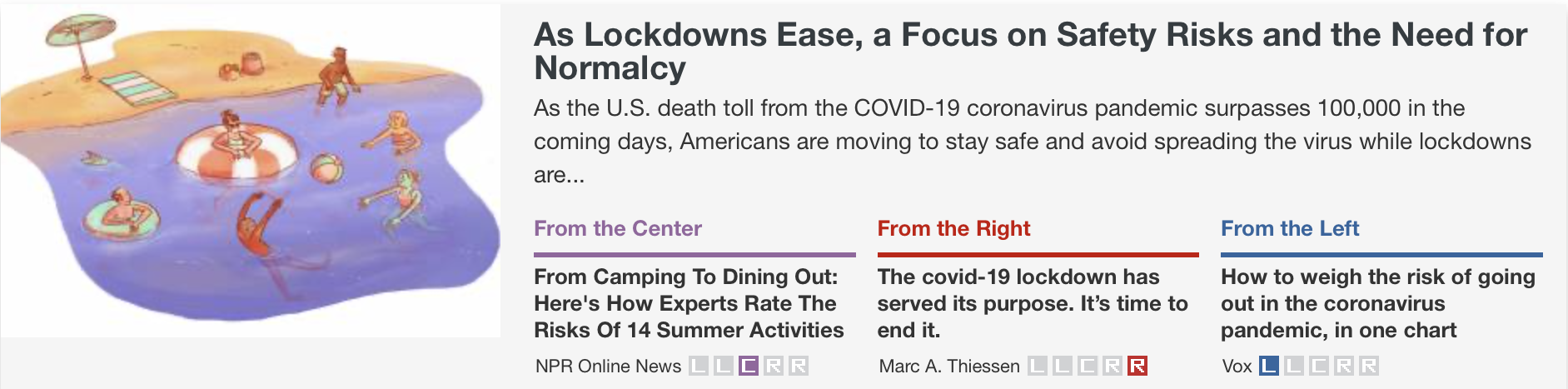}
    \caption{AllSides: balanced search on the topic of \textit{reopening after the coronavirus lockdown}.}
    \label{fig:balanced_search}
\end{figure*}


\section{Dataset}
\label{sec:dataset}
In this section, we describe the dataset that we created and that we used in our experiments. 
While most of the platforms that analyze the political leaning of news media provide in-depth analysis of particular aspects of the media, AllSides stands out as it provides annotations of political ideology for individual articles, which ensures high-quality data for both training and testing, which is in contrast with distant supervision approaches used in most previous research, as we have seen above.
In AllSides, these annotations are made as a result of a rigorous process that involves blind bias surveys, editorial reviews, third-party analysis, independent reviews, and community feedback.\footnote{\url{http://www.allsides.com/media-bias/media-bias-rating-methods}}

Furthermore, AllSides uses the annotated articles to enable its \textit{Balanced Search}, which shows news coverage on a given topic from media with different political bias.
In other words, for each trending event or topic (e.g.,~\textit{impeachment} or \textit{coronavirus pandemic}), the platform pushes news articles from all sides of the political spectrum, as shown in Figure~\ref{fig:balanced_search}.
We took advantage of this and downloaded all articles along with their political ideology annotations (\textit{left}, \textit{center}, or \textit{right}), their assigned topic(s), the media in which they were published, their author(s), and their publication date. 
Thus, our dataset contains articles that were manually selected and annotated, and that are representative of the real political scenery.
Note that the \textit{center} class covers articles that are biased towards a centrist political ideology, and not articles that lack political bias (e.g., \emph{sports} and \emph{technology}), which commonly exist in news corpora that were built by scraping RSS feeds.

We collected a total of 34,737 articles published by 73 news media and covering 109 topics.\footnote{In some cases, an article could be assigned to multiple topics, e.g.,~it could go simultaneously into \emph{coronavirus}, \emph{public health}, and \emph{healthcare}.} 
In this dataset, a total of 1,080 individual articles (3.11\%) have a political ideology label that is different from their source's.
This suggests that, while the distant supervision assumption generally holds, we would still find many articles that defy it.
Table~\ref{tbl:label_stats} shows some statistics about the dataset.

\begin{table}[h!]
    \centering
    \scalebox{0.95}{
    \begin{tabular}{lrr}
        \toprule
        \bf Political Ideology  & \bf Count & \bf Percentage\\ \midrule
        Left                    & 12,003    & 34.6\% \\
        Center                  & 9,743     & 28.1\% \\
        Right                   & 12,991    & 37.3\% \\ \bottomrule
    \end{tabular}}
    \caption{Statistics about our dataset.}
    \label{tbl:label_stats}
\end{table}

\begin{figure}[h!]
    \centering
\includegraphics[width=0.465\textwidth]{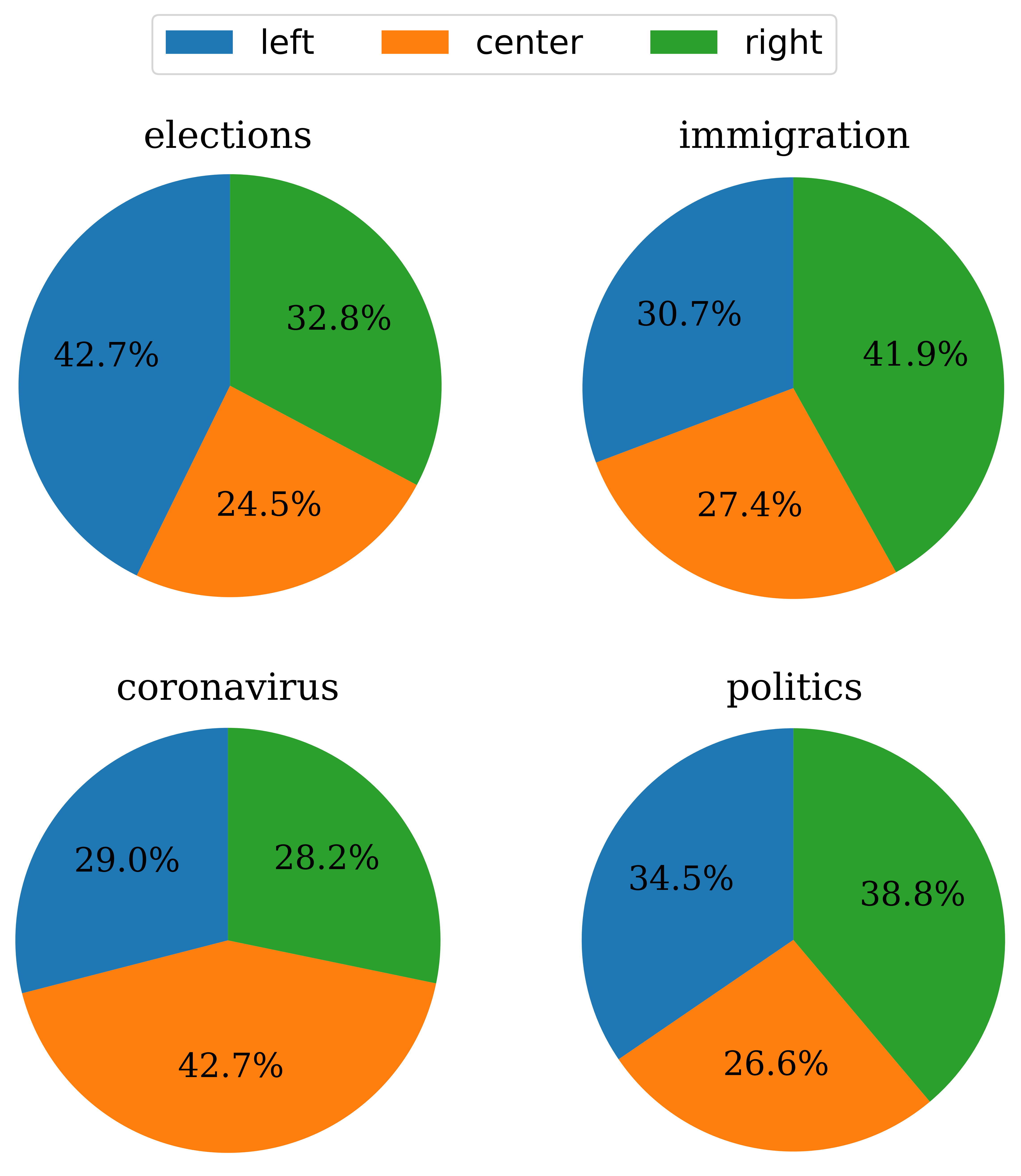}
    \caption{Political ideology for the most frequent topics: \textit{elections}, \textit{immigration}, \textit{coronavirus}, and \textit{politics}.}
    \label{fig:topics_biases}
\end{figure}

Figure~\ref{fig:topics_biases} illustrates the distribution of the different political bias labels within each of the most frequent topics.  We can see that our dataset is able to represent topics or events from different political perspectives.
This is yet another advantage, as it enables a more challenging task for machine learning models to detect the linguistic and the semantic nuances of different political ideologies in news articles, as opposed to cases where certain topics might be coincidentally collocated with certain labels, in which case the models would be actually learning to detect the topics instead of predicting the political ideology of the target news article.

\vspace{0.5cm}

It is worth noting that since most article labels are aligned with their source labels, it is likely that machine learning classifiers would end up modeling the source instead of the political ideology of the individual articles. For example, a model would be learning the writing style of each medium, and then it would associate it with a particular ideology.
Therefore, we pre-processed the articles in a way that eliminates explicit markers such as the name of the authors, or the name of the medium that usually appears as a preamble to the article's content, or in the content itself.
Furthermore, in order to ensure that we are actually modeling the political ideology as it is expressed in the language of the news, we created evaluation splits in two different ways: \Ni~randomly, which is what is typically done (for comparison only), and \Nii~based on media, where all articles by the same medium appear in either the training, the validation, or the testing dataset. 

The latter form of splitting would help us indicate what a trained classifier has actually learned.
For instance, if it modeled the source, then it would not be able to perform well on the test set, since all its articles would belong to sources that were never seen during training.
In order to ensure fair one-to-one comparisons between experiments, we created these two different sets of splits, while making sure that they share the same test set, as follows:

\begin{itemize}
    \item \textbf{Media-based Split:} We sampled 1,200 articles from 12 news media (100 per medium) and used them as the \textit{test} set, and we excluded the remaining 5,470 articles from these media.
    Then, we used the articles from the remaining 61 media to create the \textit{training} and the \textit{validation} sets, where all articles from the same medium would appear in the same set: training, development, or testing.
    This ensures that the model is fine-tuned and tested on articles whose sources were not seen during training.
    \item \textbf{Random Split:} Here, the \textit{test} set is the same as in the media-based split.
    The 5,470 articles that we excluded from the 12 media are now added to the articles from the 61 remaining media.
    Then, we split this collection of articles (using stratified random sampling) into \textit{training} and \textit{validation} sets.
    This ensures that the model is fine-tuned and evaluated only on articles whose sources were observed during training.
\end{itemize}

Table~\ref{tbl:splits_stats} shows statistics about both splits, including the size of each set and the number of media and topics they cover.
We release the dataset, along with the evaluation splits, and the code,\footnote{\url{http://github.com/ramybaly/Article-Bias-Prediction}} which can be used to extend the dataset as more news articles are added to AllSides.

\begin{table}[h!]
    \centering
    \scalebox{0.95}{
    \begin{tabular}{llrrr}
    \toprule
                                        &               & \bf Train     & \bf Valid.        & \bf Test \\ \midrule
    \multirow{3}{*}{\bf Media-based}    & \it Count     & 22,969    & 5,098         & 1,200 \\
                                        & \it Media     & 46        & 15            & 12 \\
                                        & \it Topics    & 108       & 105           & 93 \\ \midrule
    \multirow{3}{*}{\bf Random}         & \it Count     & 26,828    & 6,709         & 1,200 \\
                                        & \it Media     & 73        & 73            & 12 \\
                                        & \it Topics    & 108       & 107           & 93 \\ \bottomrule
    \end{tabular}}
    \caption{Statistics about our dataset and its two splits: \textit{media-based} and \textit{random}.}
    \label{tbl:splits_stats}
\end{table}


\section{Methodology}
\label{sec:methodology}

\subsection{Classifiers}

The task of predicting the political ideology of news articles is typically formulated as a classification problem, where the textual content of the articles is encoded into a vector representation that is used to train a classifier to predict one of $C$ classes (in our case, $C=3$: \textit{left}, \textit{center}, and \textit{right}).
In our experiments, we use two deep learning architectures: 
\Ni~\textit{Long Short-Term Memory networks} (LSTMs), which are Recurrent Neural Networks (RNNs), which use gating mechanisms to selectively pass information across time and to model long-term dependencies~\cite{hochreiter1997long}, and \Nii~\textit{Bidirectional Encoder Representations from Transformers} (BERT), with a complex architecture yielding high-quality contextualized embeddings, which have been successful in several Natural Language Processing tasks~\cite{devlin2019bert}.

\subsection{Removing Media Bias}
\label{sec:debias}

Ultimately, our goal is to develop a model that can predict the political ideology of a news article.
Our dataset, along with some others, has a special property that might stand in the way of achieving this goal.
Most articles published by a given source have the same ideological leaning.
This might confuse the model and cause it to erroneously associate the output classes with features that characterize entire media outlets (such as detecting specific writing patterns, or stylistic markers in text).
Consequently, the model would fail when applied to articles that were published in media that were unseen during training.
The experiments in Section~\ref{sec:results} confirm this.
Thus, we apply two techniques to \textit{de-bias} the models, i.e.,~to prevent them from learning the style of a specific news medium rather than predicting the political ideology of the target news article.

\subsubsection{Adversarial Adaptation (AA)}

This model was originally proposed by~\citet{ganin2016domain} for unsupervised domain adaptation in image classification.
Their objective was to adapt a model trained on labelled images from a \textit{source} domain to a novel \textit{target} domain, where the images have no labels for the task at hand.
This is done by adding an adversarial \textit{domain classifier} with a gradient reversal layer to predict the examples' domains.
The \textit{label predictor's} is minimized for the labelled examples (from the source domain), and the adversarial \textit{domain classifier's} loss is maximized for all examples in the dataset.
As a result, the encoder can extract representation that is \Ni~discriminative for the main task and also \Nii~invariant across domains (due to the gradient reversal layer).
The overall loss is minimized as follows:

\begin{equation}
    \sum_{\substack{i=1:N \\ d_i=0}} \mathcal{L}_y^i(\theta_f, \theta_y) - \lambda\sum_{i=1:N}\mathcal{L}^i_d(\theta_f, \theta_d),\label{eq:aa_1}
\end{equation}

\noindent where $N$ is the number of training examples, $\mathcal{L}^i_y(\cdot,\cdot)$ is the label predictor's loss, the condition $d_i=0$ means that only examples from the source domain are used to calculate the label predictor's loss, $\mathcal{L}^i_d(\cdot,\cdot)$ is the domain classifier's loss, $\lambda$ controls the trade-off between both losses, and $\{\theta_f,\theta_y,\theta_d\}$ are the parameters of the encoder, the label predictor, and the domain classifier, respectively.
Further details about the formulation of this method is available in~\cite{ganin2016domain}.

We adapt this architecture as follows.
Instead of a \textit{domain classifier}, we implement a \textit{media classifier}, which, given an article, tries to predict the medium it comes from.
As a result, the encoder should extract representation that is discriminative for the main task of predicting political ideology, while being invariant for the different media. 
This approach was originally proposed as an unsupervised domain adaptation, since labelled examples were available for one domain only, whereas in our case, all articles from different media were labelled for their political ideology.
Therefore, we jointly minimize the losses of both the \textit{label predictor} and the \textit{media classifier} over the entire dataset.
The new objective function to minimize is as follows:

\begin{equation}
    \sum_{i=1:N} \mathcal{L}_y^i(\theta_f, \theta_y) - \lambda\sum_{i=1:N}\mathcal{L}^i_m(\theta_f, \theta_m),\label{eq:aa_2}
\end{equation}

\noindent where $\mathcal{L}^i_m(\cdot,\cdot)$ is the loss of the \textit{media classifier}, and $\theta_m$ is its set of parameters.

\subsubsection{Triplet Loss Pre-training (TLP)}

In this approach, we pre-train the encoder using a triplet loss~\cite{schroff2015facenet}.
The model is trained on a set of triplets, each composed of an anchor, a positive, and a negative example.
The objective in Eq.~\ref{eq:triplet_loss} ensures that the positive example is always closer to the anchor than the negative example is, where $\boldsymbol{a}$, $\boldsymbol{p}$ and $\boldsymbol{n}$ are the encodings of the anchor, of the positive, and of the negative examples, respectively, and $D(\cdot, \cdot)$ is the Euclidean distance:
\begin{equation}
    \mathcal{L} =\max\left(D\left(\boldsymbol{a},\boldsymbol{p}\right) - D\left(\boldsymbol{a},\boldsymbol{n}\right) + \epsilon, 0\right).\label{eq:triplet_loss}
\end{equation}

\begin{figure}[h!t]
    \centering
    \includegraphics[width=0.475\textwidth]{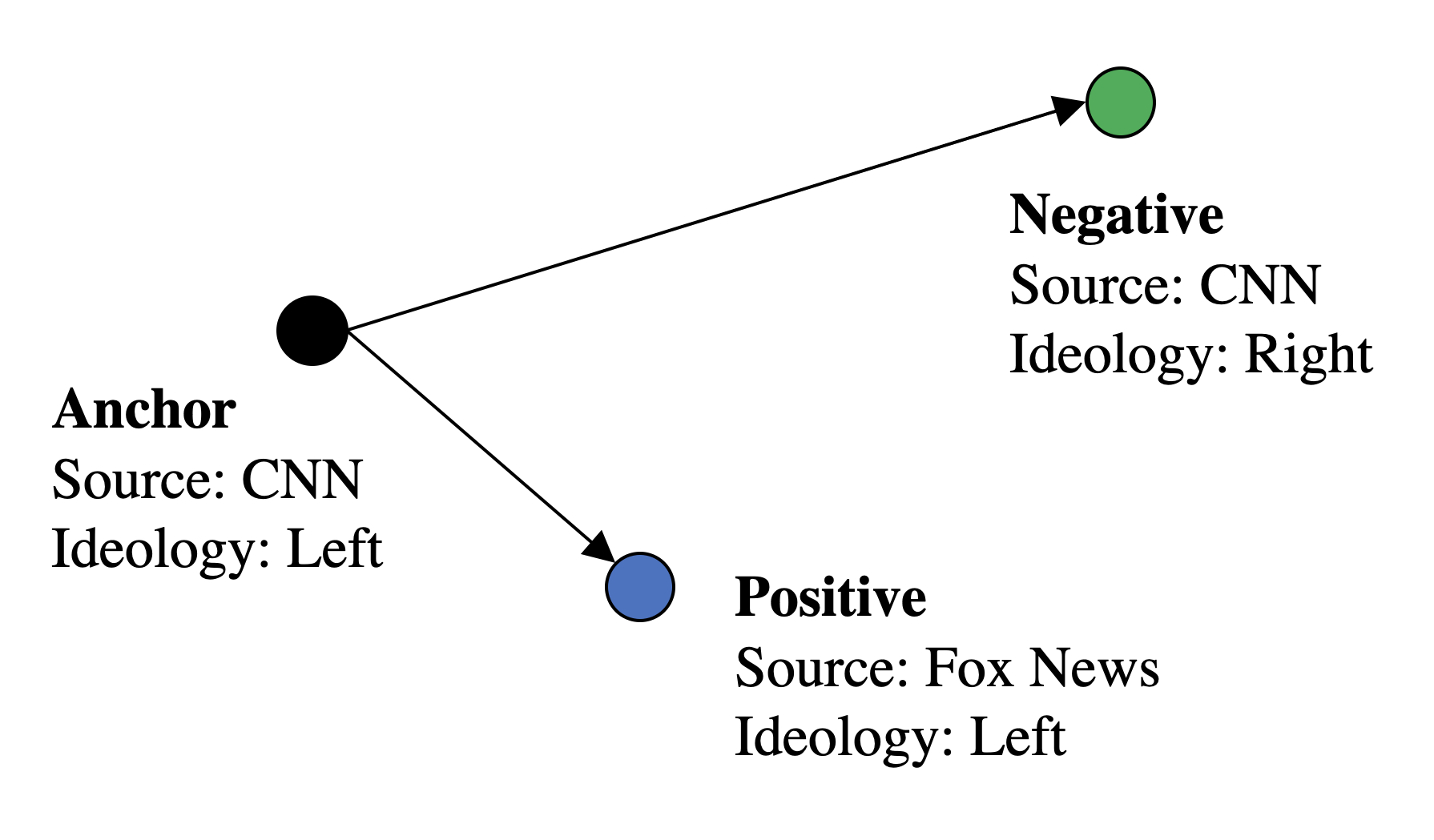}
    \caption{An example triplet used for de-biasing.}
    \label{fig:adaptationtriplet}
\end{figure}

Figure~\ref{fig:adaptationtriplet} shows an example of such a triplet.
The positive example shares the same ideology as the anchor's, but they are published by different media.
The negative example has a different ideology than the anchor's, but they are published by the same medium.
In this way, the encoder will be clustering examples with similar ideologies close to each other, regardless of their source.
Once the encoder has been pre-trained, its parameters, along with the softmax classifier's, are fine-tuned on the main task by minimizing the cross-entropy loss when predicting the political ideology of articles.

\subsection{Media-level Representation}
\label{sec:media_features}

Finally, we explore the benefits of incorporating information describing the target medium, which can serve as a complementary representation for the article.
While this seems to be counter-intuitive to what we have been proposing in Subsection~\ref{sec:debias}, we believe that medium-level representation can be valuable when combined with an accurate representation of the article.
Intuitively, having an accurate understanding of the natural language in the article, together with a glimpse into the medium it is published in, should provide a more complete picture of its underlying political ideology.

\citet{baly2020written} proposed a comprehensive set of representation to characterize news media from different angles: how a medium portrays itself, who is its audience, and what is written about it.
Their results indicate that exploring the \textit{Twitter bios} of a medium's followers offers a good insight into its political leaning.
To a lesser extent, the content of a \textit{Wikipedia} page describing a medium can also help unravel its political leaning.
Therefore, we concatenated these representations to the encoded articles, at the output of the encoder and right before the \textsc{softmax} layer, so that both the article encoder and the classification layer that is based on the article and the external media representations are trained jointly and end-to-end.

Similarly to~\cite{baly2020written}, we retrieved the profiles of up to a 1,000 Twitter followers for each medium, we encoded their bios using the Sentence-BERT model~\cite{reimers2019sentence}, and we then averaged these encodings to obtain a single representation for that medium.
As for the Wikipedia representation, we automatically retrieved the content of the page describing each medium, whenever applicable.
Then, we used the pre-trained base BERT model to encode this content by averaging the word representations extracted from BERT's second-to-last layer, which is common practice, since the last layer may be biased towards the pre-training objectives of BERT.


\section{Experiments and Results}
\label{sec:results}

We evaluated both the LSTM and the BERT models, assessing the impact of \Ni~de-biasing and \Nii~incorporating media-level representation.

\subsection{Experimental Setup}

We fine-tuned the hyper-parameters of both models on the validation set using a guided grid search trial while fixing the seeds of the random weights initialization.
For LSTM, we varied the length of the input (128--1,024 tokens), the number of layers (1--3), the size of the LSTM cell (200--400), the dropout rate (0--0.8), the learning rate ($1\mathrm{e}{-3}$ to $1\mathrm{e}{-5}$), the gradient clipping value (0--5), and the batch size (8--256).
The best results were obtained with a 512-token input, a 2-layer LSTM of size 256, a dropout rate of 0.7, a learning rate of $1\mathrm{e}{-3}$, gradient clipping at 0.5, and a batch size of 32.
This model has around 1.1M trainable parameters, and was trained with 300-dimensional GloVe input word embeddings~\cite{pennington2014glove}.

For BERT, we varied the length of the input, the learning rate, and the gradient clipping value.
The best results were obtained using a 512-token input, a learning rate of $2\mathrm{e}{-5}$, and gradient clipping at 1.
This model has 110M trainable parameters.

We trained our models on 4 \textit{Titan X Pascal} GPUs, and the runtime for each epoch was 25 seconds for the LSTM-based models and 22 minutes for the BERT-based models.
For each experiment, the model was trained only once with fixed seeds used to initialize the models' weights.

For the Adversarial Adaptation (AA), we have an additional hyper-parameter $\lambda$ (see Equation~\ref{eq:aa_2}), which we varied from 0 to 1, where 0 means no adaptation at all.
The best results were obtained with $\lambda=0.7$, which means that we need to pay significant attention to the adversarial classifier's loss in order to mitigate the media bias.

For the Triplet Loss Pre-training (PLT), we sampled 35,017 triplets from the training set, such that the examples in each triplet discuss the same topic in order to ensure that the change in topic has minimal impact on the distance between the examples.

To evaluate our models, we use accuracy and macro-$F_1$ score ($F_1$ averaged across all classes), which we also used as an early stopping criterion, since the classes were slightly imbalanced.
Moreover, given the ordinal nature of the labels, we report the Mean Absolute Error (MAE), shown in Equation~\eqref{eq:mae}, where $N$ is the number of instances, and $y_i$ and $\hat{y}_i$ are the number of correct and of predicted labels, respectively.

\begin{equation}
    \text{MAE} = \frac{1}{N}\sum_{i=1}^N\left|y_i-\hat{y}_i\right|\label{eq:mae}
\end{equation}

\subsection{Results}

\paragraph{Baseline Results}
The results in Table~\ref{tbl:baselines} show the performance for LSTM and for BERT at predicting the political ideology of news articles for both the \textit{media-based} and the \textit{random} splits.
We observe sizable differences in performance between the two splits.
In particular, both models perform much better when they are trained and evaluated on the \textit{random} split, whereas they both fail on the \textit{media-based} split, where they are tested on articles from media that were not seen during training.
This observation confirms our initial concerns that the models would tend to learn general characteristics about news media, and then would face difficulties with articles coming from new unseen media.

\begin{table}[h!]
    \centering
    \scalebox{0.85}{
    \begin{tabular}{llccc}
        \toprule
        \bf Model                   & \bf Split         & \bf Macro $\boldsymbol{F_1}$  & \bf Acc.  & \bf MAE \\ \midrule
        \it Majority                &                   & 19.61                         & 41.67     & 0.92 \\ \midrule
        \multirow{2}{*}{\it LSTM}   & \it Media-based   & 31.51                         & 32.30     & 0.97 \\
                                    & \it Random        & 65.50                         & 66.17     & 0.52 \\ \midrule
        \multirow{2}{*}{\it BERT}   & \it Media-based   & 35.53                         & 36.75     & 0.90 \\
                                    & \it Random        & 80.19                         & 79.83     & 0.33 \\ \bottomrule
    \end{tabular}}
    \caption{Baseline experiments (without de-biasing or media-level representation) for the two splits.}
    \label{tbl:baselines}
\end{table}

\paragraph{Removing the Source Bias}
In order to further confirm the bias towards modeling the media, we ran a side experiment of fine-tuning BERT on the task of predicting the medium given the article's content, which is a 73-way classification problem.
We used stratified random sampling to create the evaluation splits and to make sure each set contains all labels (media).
The results in Table~\ref{tbl:source_prediction} confirm that BERT is much stronger than the majority class baseline, despite the high number of classes, which means that predicting the medium in which a target news article was published is a fairly easy task.

\begin{table}[h!]
    \centering
    \scalebox{1}{
    \begin{tabular}{lrr}
        \toprule
        \bf Model       & \bf Macro $\boldsymbol{F_1}$  & \bf Acc. \\ \midrule
        \it Majority    & 0.25                          & 10.21 \\
        \it BERT        & 59.72                         & 80.12 \\ \bottomrule
    \end{tabular}}
    \caption{Predicting the medium in which a target news article was published.}
    \label{tbl:source_prediction}
\end{table}

\begin{table}[b]
    \centering
    \scalebox{0.95}{
    \begin{tabular}{lcccc}
        \toprule
        \bf Model                   & \bf De-bias   & \bf Macro $\boldsymbol{F_1}$  & \bf Acc.      & \bf MAE \\ \midrule
        \multirow{3}{*}{\it LSTM}   & \it None      & 31.51                         & 32.30         & 0.97 \\
                                    & \it AA        & 40.33                         & 40.57         & 0.69 \\
                                    & \it TLP       & \bf 45.44                     & \bf 46.42     & \bf 0.62 \\ \midrule
        \multirow{3}{*}{\it BERT}   & \it None      & 35.53                         & 36.75         & 0.90 \\
                                    & \it AA        & 43.87                         & 46.22         & 0.59 \\
                                    & \it TLP       & \bf 48.26                     & \bf 51.41     & \bf 0.51 \\ \bottomrule
    \end{tabular}}
    \caption{Impact of de-biasing (adversarial adaptation and triplet loss) on article-level bias detection.}
    \label{tbl:debiasing}
\end{table}

\begin{table*}[tbh]
    \centering
    \scalebox{1}{
    \begin{tabular}{clccc@{ }@{ }@{ }@{ }@{ }ccc}
        \toprule
        \multicolumn{2}{c}{}                               & \multicolumn{3}{c}{\bf LSTM}                         & \multicolumn{3}{c}{\bf BERT} \\
        \bf \#  & \bf Representation                              & \bf Macro $\boldsymbol{F_1}$  & \bf Acc.  & \bf MAE   & \bf Macro $\boldsymbol{F_1}$  & \bf Acc.  & \bf MAE \\ \midrule
        1       & \it Article (baseline)                    & 31.51                         & 32.30     & 0.97      & 35.53                         & 36.75     & 0.90 \\
        2       & \it Article with TLP                      & 45.44                         & 46.42     & 0.62      & 48.26                         & 51.41     & 0.51 \\ \cdashlinelr{1-8}
        3       & \it Wikipedia                             & 41.39                         & 41.86     & 0.92      & 41.39                         & 41.86     & 0.92 \\ 
        4       & \it Wikipedia $+$ Article                 & 40.49                         & 40.79     & 0.92      & 42.33                         & 41.90     & 0.90 \\
        5       & \it Wikipedia $+$ Article with TLP        & 48.25                         & 46.47     & 0.69      & 51.16                         & 49.75     & 0.32 \\
        6       & \it Twitter bios                          & 60.30                         & 62.69     & 0.42      & 60.30                         & 62.69     & 0.42 \\
        7       & \it Twitter bios $+$ Article              & 60.30                         & 62.69     & 0.42      & 60.42                         & 63.12     & 0.40 \\
        8       & \it Twitter bios $+$ Article with TLP     & \bf 62.02                     & \bf 70.03 & \bf 0.32  & \bf 64.29                     & \bf 72.00 & \bf 0.29 \\ \bottomrule
    \end{tabular}}
    \caption{Impact of adding media-level representations to the article-level representations (with and without de-biasing). Note that the results in rows 3 and 6 are the same for both LSTM and BERT because no articles were involved, and the media-level representations were directly used to train the classifier.}
    \label{tbl:media_features}
\end{table*}

In order to remove the bias towards modeling the medium, we evaluated the impact of the adversarial adaptation (AA) and the Triplet Loss Pre-training (TLP) with the media-based split.
The results in Table~\ref{tbl:debiasing} show sizeable improvements when either of these approaches is used, compared to the baseline (no de-biasing). In particular, TLP yields an improvement of 14.12 points absolute in terms of accuracy, and 12.73 points in terms of macro-$F_1$.

\paragraph{Impact of Media-Level Representation}
Finally, we evaluated the impact of incorporating the media-level representation (Twitter followers' bios and Wikipedia content) in addition to teh article-level representation.
Table~\ref{tbl:media_features} illustrates these results in an incremental way.
First, we evaluated the performance of the media-level representation alone at predicting the political ideology of news articles (see rows 3 and 6).
We should note that these results are identical for the LSTM and the BERT columns since no article was encoded in these experiments, and the media representation was used directly to train the logistic regression classifier.
Then, adding the article representation from either model, without any de-biasing, had no or little impact on the performance (see rows 4~vs.~3, and 7~vs.~6).
This is not surprising, since we have shown that, without de-biasing, both models learn more about the source than about the bias in the language used by the article.
Therefore, the ill-encoded articles do not provide more information than what the medium representation already gives, which is why no or too little improvement was observed.

When we use the triplet loss to mitigate the source bias, the resulting article representation is more accurate and meaningful, and the medium representation does offer complementary information, and eventually contributes to sizeable performance gains (see rows 5 and 8~vs.~2).
The Twitter bios representation appears to be much more important than the representation from Wikipedia, which shows the importance of inspecting the media followers' background and their point of views, which is also one of the observations in~\cite{baly2020written}.

Overall, comparing the best results to the baseline (rows 8~vs.~1), we can see that \Ni~using the triplet loss to remove the source bias, and \Nii~incorporating media-level representation from Twitter followers yields 30.51 and 28.76 absolute improvement in terms of macro $\boldsymbol{F_1}$ on the challenging \textit{media-based} split.


\section{Conclusion and Future Work}
\label{sec:conclusion}

We have explored the task of predicting the leading political ideology of news articles.
In particular, we created a new large dataset for this task, which features article-level annotations and is well-balanced across topics and media.
We further proposed an adversarial media adaptation approach, as well as a special triplet loss in order to prevent modeling the source instead of the political bias in the news article, which is a common pitfall for approaches dealing with data that exhibit high correlation between the source of a news article and its class, as is the case with our task here. Finally, our experimental results have shown very sizable improvements over using state-of-the-art pre-trained Transformers.

In future work, we plan to explore topic-level bias prediction as well as going beyond left-center-right bias. We further want to develop models that would be able to detect specific fragments in an article where the bias occurs, thus enabling explainability.
Last but not least, we plan to experiment with other languages, and to explore to what extent a model for one language is transferable to another one given that the left-center-right division is not universal and does not align perfectly across countries and cultures, even when staying within the Western political world.

\section*{Acknowledgments}
This research is part of the Tanbih project\footnote{\url{http://tanbih.qcri.org/}}, which aims to limit the effect of ``fake news,'' propaganda and media bias by making users aware of what they are reading.
The project is developed in collaboration between the Qatar Computing Research Institute, HBKU and the MIT Computer Science and Artificial Intelligence Laboratory.

\bibliography{references}

\begin{thebibliography}{35}
\expandafter\ifx\csname natexlab\endcsname\relax\def\natexlab#1{#1}\fi

\bibitem[{Baly et~al.(2018)Baly, Karadzhov, Alexandrov, Glass, and
  Nakov}]{baly2018predicting}
Ramy Baly, Georgi Karadzhov, Dimitar Alexandrov, James Glass, and Preslav
  Nakov. 2018.
\newblock Predicting factuality of reporting and bias of news media sources.
\newblock In \emph{Proceedings of the 2018 Conference on Empirical Methods in
  Natural Language Processing}, EMNLP~'18, pages 3528--3539, Brussels, Belgium.

\bibitem[{Baly et~al.(2020)Baly, Karadzhov, An, Kwak, Dinkov, Ali, Glass, and
  Nakov}]{baly2020written}
Ramy Baly, Georgi Karadzhov, Jisun An, Haewoon Kwak, Yoan Dinkov, Ahmed Ali,
  James Glass, and Preslav Nakov. 2020.
\newblock What was written vs. who read it: News media profiling using text
  analysis and social media context.
\newblock In \emph{Proceedings of the 58th Annual Meeting of the Association
  for Computational Linguistics}, ACL~'20, pages 3364--3374.

\bibitem[{Baly et~al.(2019)Baly, Karadzhov, Saleh, Glass, and
  Nakov}]{baly2019multi}
Ramy Baly, Georgi Karadzhov, Abdelrhman Saleh, James Glass, and Preslav Nakov.
  2019.
\newblock Multi-task ordinal regression for jointly predicting the
  trustworthiness and the leading political ideology of news media.
\newblock In \emph{Proceedings of the 17th Annual Conference of the North
  American Chapter of the Association for Computational Linguistics: Human
  Language Technologies}, NAACL-HLT~'19, pages 2109--2116, Minneapolis, MN,
  USA.

\bibitem[{Barr{\'o}n-Cedeno et~al.(2019)Barr{\'o}n-Cedeno, Jaradat,
  Da~San~Martino, and Nakov}]{BARRONCEDENO20191849}
Alberto Barr{\'o}n-Cedeno, Israa Jaradat, Giovanni Da~San~Martino, and Preslav
  Nakov. 2019.
\newblock {Proppy: Organizing the news based on their propagandistic content}.
\newblock \emph{Information Processing \& Management}, 56(5):1849--1864.

\bibitem[{Budak et~al.(2016)Budak, Goel, and Rao}]{budak2016fair}
Ceren Budak, Sharad Goel, and Justin~M Rao. 2016.
\newblock Fair and balanced? {Q}uantifying media bias through crowdsourced
  content analysis.
\newblock \emph{Public Opinion Quarterly}, 80(S1):250--271.

\bibitem[{Da~San~Martino et~al.(2020{\natexlab{a}})Da~San~Martino,
  Barr\'{o}n-Cede\~no, Wachsmuth, Petrov, and
  Nakov}]{DaSanMartinoSemeval20task11}
Giovanni Da~San~Martino, Alberto Barr\'{o}n-Cede\~no, Henning Wachsmuth,
  Rostislav Petrov, and Preslav Nakov. 2020{\natexlab{a}}.
\newblock {SemEval}-2020 task 11: Detection of propaganda techniques in news
  articles.
\newblock In \emph{Proceedings of the International Workshop on Semantic
  Evaluation}, SemEval~'20, Barcelona, Spain.

\bibitem[{Da~San~Martino et~al.(2020{\natexlab{b}})Da~San~Martino, Cresci,
  Barr\'{o}n-Cede\~no, Yu, Di~Pietro, and Nakov}]{IJCAI2020:propaganda:survey}
Giovanni Da~San~Martino, Stefano Cresci, Alberto Barr\'{o}n-Cede\~no, Seunghak
  Yu, Roberto Di~Pietro, and Preslav Nakov. 2020{\natexlab{b}}.
\newblock A survey on computational propaganda detection.
\newblock In \emph{Proceedings of the 29th International Joint Conference on
  Artificial Intelligence and the 17th Pacific Rim International Conference on
  Artificial Intelligence}, IJCAI-PRICAI~'20, pages 4826--4832, Yokohama,
  Japan.

\bibitem[{Da~San~Martino et~al.(2019)Da~San~Martino, Yu, Barron-Cedeno, Petrov,
  and Nakov}]{EMNLP2019:propaganda:finegrained}
Giovanni Da~San~Martino, Seunghak Yu, Alberto Barron-Cedeno, Rostislav Petrov,
  and Preslav Nakov. 2019.
\newblock Fine-grained analysis of propaganda in news articles.
\newblock In \emph{Proceedings of the 2019 Conference on Empirical Methods in
  Natural Language Processing}, EMNLP~'19, pages 5636--5646, Hong Kong, China.

\bibitem[{Darwish et~al.(2020)Darwish, Aupetit, Stefanov, and
  Nakov}]{ICWSM2020:Unsupervised:Stance:Twitter}
Kareem Darwish, Michael Aupetit, Peter Stefanov, and Preslav Nakov. 2020.
\newblock Unsupervised user stance detection on {T}witter.
\newblock In \emph{Proceedings of the International AAAI Conference on Web and
  Social Media}, ICWSM~'20, pages 141--152, Atlanta, GA, USA.

\bibitem[{DellaVigna and Kaplan(2007)}]{dellavigna2007fox}
Stefano DellaVigna and Ethan Kaplan. 2007.
\newblock The {Fox News} effect: Media bias and voting.
\newblock \emph{The Quarterly Journal of Economics}, 122(3):1187--1234.

\bibitem[{Devlin et~al.(2019)Devlin, Chang, Lee, and
  Toutanova}]{devlin2019bert}
Jacob Devlin, Ming-Wei Chang, Kenton Lee, and Kristina Toutanova. 2019.
\newblock {BERT}: Pre-training of deep bidirectional transformers for language
  understanding.
\newblock In \emph{Proceedings of the 2019 Conference of the North American
  Chapter of the Association for Computational Linguistics: Human Language
  Technologies}, NAACL-HLT~'19, pages 4171--4186, Minneapolis, MN, USA.

\bibitem[{Dinkov et~al.(2019)Dinkov, Ali, Koychev, and
  Nakov}]{DBLP:conf/interspeech/Dinkov0KN19}
Yoan Dinkov, Ahmed Ali, Ivan Koychev, and Preslav Nakov. 2019.
\newblock Predicting the leading political ideology of {YouTube} channels using
  acoustic, textual, and metadata information.
\newblock In \emph{Proceedings of the 20th Annual Conference of the
  International Speech Communication Association}, INTERSPEECH~'19, pages
  501--505, Graz, Austria.

\bibitem[{Elejalde et~al.(2018)Elejalde, Ferres, and
  Herder}]{elejalde2018nature}
Erick Elejalde, Leo Ferres, and Eelco Herder. 2018.
\newblock {On the nature of real and perceived bias in the mainstream media}.
\newblock \emph{PloS one}, 13(3):e0193765.

\bibitem[{Ganin et~al.(2016)Ganin, Ustinova, Ajakan, Germain, Larochelle,
  Laviolette, Marchand, and Lempitsky}]{ganin2016domain}
Yaroslav Ganin, Evgeniya Ustinova, Hana Ajakan, Pascal Germain, Hugo
  Larochelle, Fran{\c{c}}ois Laviolette, Mario Marchand, and Victor Lempitsky.
  2016.
\newblock {Domain-adversarial training of neural networks}.
\newblock \emph{The Journal of Machine Learning Research}, 17(1):2096--2030.

\bibitem[{Graber and Dunaway(2017)}]{graber2017mass}
Doris~A Graber and Johanna Dunaway. 2017.
\newblock \emph{Mass media and American politics}.
\newblock SAGE Publications.

\bibitem[{Gruppi et~al.(2020)Gruppi, Horne, and Adalı}]{gruppi2020nelagt2019}
Maurício Gruppi, Benjamin~D. Horne, and Sibel Adalı. 2020.
\newblock {NELA-GT-2019}: A large multi-labelled news dataset for the study of
  misinformation in news articles.
\newblock \emph{arXiv preprint arXiv:2003.08444}.

\bibitem[{Hochreiter and Schmidhuber(1997)}]{hochreiter1997long}
Sepp Hochreiter and J{\"u}rgen Schmidhuber. 1997.
\newblock {Long Short-Term Memory}.
\newblock \emph{Neural Computation}, 9(8):1735--1780.

\bibitem[{Horne et~al.(2018)Horne, Dron, Khedr, and
  Adali}]{Horne:2018:ANL:3184558.3186987}
Benjamin~D. Horne, William Dron, Sara Khedr, and Sibel Adali. 2018.
\newblock Assessing the news landscape: A multi-module toolkit for evaluating
  the credibility of news.
\newblock In \emph{Proceedings of the The Web Conference}, WWW~'18, pages
  235--238, Lyon, France.

\bibitem[{Horne et~al.(2019)Horne, N{\o}rregaard, and
  Adal{\i}}]{horne2019different}
Benjamin~D Horne, Jeppe N{\o}rregaard, and Sibel Adal{\i}. 2019.
\newblock Different spirals of sameness: A study of content sharing in
  mainstream and alternative media.
\newblock In \emph{Proceedings of the International AAAI Conference on Web and
  Social Media}, ICWSM~'19, pages 257--266, Munich, Germany.

\bibitem[{Iyengar and Hahn(2009)}]{iyengar2009red}
Shanto Iyengar and Kyu~S Hahn. 2009.
\newblock Red media, blue media: Evidence of ideological selectivity in media
  use.
\newblock \emph{Journal of communication}, 59(1):19--39.

\bibitem[{Jiang et~al.(2019)Jiang, Petrak, Song, Bontcheva, and
  Maynard}]{jiang2019team}
Ye~Jiang, Johann Petrak, Xingyi Song, Kalina Bontcheva, and Diana Maynard.
  2019.
\newblock {Team Bertha von Suttner at SemEval-2019 Task 4}: Hyperpartisan news
  detection using {ELMo} sentence representation convolutional network.
\newblock In \emph{Proceedings of the 13th International Workshop on Semantic
  Evaluation}, SemEval~'19, pages 840--844, Minneapolis, MN, USA.

\bibitem[{Kiesel et~al.(2019)Kiesel, Mestre, Shukla, Vincent, Adineh, Corney,
  Stein, and Potthast}]{kiesel-etal-2019-semeval}
Johannes Kiesel, Maria Mestre, Rishabh Shukla, Emmanuel Vincent, Payam Adineh,
  David Corney, Benno Stein, and Martin Potthast. 2019.
\newblock {SemEval-2019 Task 4}: Hyperpartisan news detection.
\newblock In \emph{Proceedings of the 13th International Workshop on Semantic
  Evaluation}, SemEval~'19, pages 829--839, Minneapolis, Minnesota, USA.

\bibitem[{Kulkarni et~al.(2018)Kulkarni, Ye, Skiena, and
  Wang}]{kulkarni2018multi}
Vivek Kulkarni, Junting Ye, Steven Skiena, and William~Yang Wang. 2018.
\newblock Multi-view models for political ideology detection of news articles.
\newblock In \emph{Proceedings of the Conference on Empirical Methods in
  Natural Language Processing}, EMNLP~'18, pages 3518--3527, Brussels, Belgium.

\bibitem[{Lin et~al.(2006)Lin, Wilson, Wiebe, and
  Hauptmann}]{lin-etal-2006-side}
Wei-Hao Lin, Theresa Wilson, Janyce Wiebe, and Alexander Hauptmann. 2006.
\newblock Which side are you on? {I}dentifying perspectives at the document and
  sentence levels.
\newblock In \emph{Proceedings of the Tenth Conference on Computational Natural
  Language Learning}, {C}o{NLL}~'06, pages 109--116.

\bibitem[{Pang and Lee(2004)}]{pang-lee-2004-sentimental}
Bo~Pang and Lillian Lee. 2004.
\newblock A sentimental education: Sentiment analysis using subjectivity
  summarization based on minimum cuts.
\newblock In \emph{Proceedings of the 42nd Annual Meeting of the Association
  for Computational Linguistics}, ACL~'04, pages 271--278, Barcelona, Spain.

\bibitem[{Pennington et~al.(2014)Pennington, Socher, and
  Manning}]{pennington2014glove}
Jeffrey Pennington, Richard Socher, and Christopher~D Manning. 2014.
\newblock {GloVe}: Global vectors for word representation.
\newblock In \emph{Proceedings of the 2014 Conference on Empirical Methods in
  Natural Language Processing}, EMNLP~'14, pages 1532--1543, Doha, Qatar.

\bibitem[{Potthast et~al.(2018)Potthast, Kiesel, Reinartz, Bevendorff, and
  Stein}]{DBLP:journals/corr/PotthastKRBS17}
Martin Potthast, Johannes Kiesel, Kevin Reinartz, Janek Bevendorff, and Benno
  Stein. 2018.
\newblock A stylometric inquiry into hyperpartisan and fake news.
\newblock In \emph{Proceedings of the 56th Annual Meeting of the Association
  for Computational Linguistics}, ACL~'18, pages 231--240, Melbourne,
  Australia.

\bibitem[{Rashkin et~al.(2017)Rashkin, Choi, Jang, Volkova, Choi, and
  Allen}]{Rashkin}
Hannah Rashkin, Eunsol Choi, Jin~Yea Jang, Svitlana Volkova, Yejin Choi, and
  Paul~G Allen. 2017.
\newblock Truth of varying shades: Analyzing language in fake news and
  political fact-checking.
\newblock In \emph{Proceedings of the 2017 Conference on Empirical Methods in
  Natural Language Processing}, EMNLP~'17, pages 2931--2937, Copenhagen,
  Denmark.

\bibitem[{Reimers and Gurevych(2019)}]{reimers2019sentence}
Nils Reimers and Iryna Gurevych. 2019.
\newblock {Sentence-BERT}: Sentence embeddings using {S}iamese {BERT}-networks.
\newblock In \emph{Proceedings of the 2019 Conference on Empirical Methods in
  Natural Language Processing and the 9th International Joint Conference on
  Natural Language Processing}, EMNLP-IJCNLP~'19, pages 3973--3983, Hong Kong,
  China.

\bibitem[{Saez-Trumper et~al.(2013)Saez-Trumper, Castillo, and
  Lalmas}]{10.1145/2505515.2505623}
Diego Saez-Trumper, Carlos Castillo, and Mounia Lalmas. 2013.
\newblock Social media news communities: Gatekeeping, coverage, and statement
  bias.
\newblock In \emph{Proceedings of the 22nd ACM International Conference on
  Information \& Knowledge Management}, CIKM~'13, page 1679–1684, San
  Francisco, CA, USA.

\bibitem[{Saleh et~al.(2019)Saleh, Baly, Barr{\'o}n-Cede{\~n}o, Da~San~Martino,
  Mohtarami, Nakov, and Glass}]{saleh2019team}
Abdelrhman Saleh, Ramy Baly, Alberto Barr{\'o}n-Cede{\~n}o, Giovanni
  Da~San~Martino, Mitra Mohtarami, Preslav Nakov, and James Glass. 2019.
\newblock {Team QCRI-MIT at SemEval-2019 Task 4}: Propaganda analysis meets
  hyperpartisan news detection.
\newblock In \emph{Proceedings of the 13th International Workshop on Semantic
  Evaluation}, SemEval~'19, pages 1041--1046, Minneapolis, MN, USA.

\bibitem[{Schroff et~al.(2015)Schroff, Kalenichenko, and
  Philbin}]{schroff2015facenet}
Florian Schroff, Dmitry Kalenichenko, and James Philbin. 2015.
\newblock {FaceNet}: A unified embedding for face recognition and clustering.
\newblock In \emph{Proceedings of the IEEE Conference on Computer Vision and
  Pattern Recognition}, CVPR~'15, pages 815--823, Boston, MA, USA.

\bibitem[{Sim et~al.(2013)Sim, Acree, Gross, and
  Smith}]{sim-etal-2013-measuring}
Yanchuan Sim, Brice D.~L. Acree, Justin~H. Gross, and Noah~A. Smith. 2013.
\newblock Measuring ideological proportions in political speeches.
\newblock In \emph{Proceedings of the 2013 Conference on Empirical Methods in
  Natural Language Processing}, EMNLP~'13, pages 91--101, Seattle, Washington,
  USA.

\bibitem[{Stefanov et~al.(2020)Stefanov, Darwish, Atanasov, and
  Nakov}]{stefanov-etal-2020-predicting}
Peter Stefanov, Kareem Darwish, Atanas Atanasov, and Preslav Nakov. 2020.
\newblock Predicting the topical stance and political leaning of media using
  tweets.
\newblock In \emph{Proceedings of the 58th Annual Meeting of the Association
  for Computational Linguistics}, ACL~'20, pages 527--537.

\bibitem[{Zhang et~al.(2019)Zhang, Da~San~Martino, Barr{\'o}n-Cede{\~n}o,
  Romeo, An, Kwak, Staykovski, Jaradat, Karadzhov, Baly, Darwish, Glass, and
  Nakov}]{zhang-etal-2019-tanbih}
Yifan Zhang, Giovanni Da~San~Martino, Alberto Barr{\'o}n-Cede{\~n}o, Salvatore
  Romeo, Jisun An, Haewoon Kwak, Todor Staykovski, Israa Jaradat, Georgi
  Karadzhov, Ramy Baly, Kareem Darwish, James Glass, and Preslav Nakov. 2019.
\newblock {T}anbih: Get to know what you are reading.
\newblock In \emph{Proceedings of the 2019 Conference on Empirical Methods in
  Natural Language Processing and the 9th International Joint Conference on
  Natural Language Processing}, EMNLP-IJCNLP~'19, pages 223--228, Hong Kong,
  China.

\end{thebibliography}
\bibliographystyle{acl_natbib}

\end{document}